\documentclass[preprint,review,12pt]{elsarticle}

\usepackage{textcomp}  
\usepackage{amsmath,amsfonts,amssymb}
\usepackage{amsthm}
\usepackage{booktabs}
\usepackage{subfigure}
\usepackage{txfonts} 
\usepackage[linesnumbered,ruled,vlined]{algorithm2e}

\journal{}

\begin{document}
\begin{frontmatter}


\title{Root-KGD: A Novel Framework for Root Cause Diagnosis Based on Knowledge Graph and Industrial Data}






\author{Jiyu Chen}
\ead{22332128@zju.edu.cn}

\author{Jinchuan Qian}
\ead{qianjinchuan@zju.edu.cn}

\author{Xinmin Zhang*}
\ead{xinminzhang@zju.edu.cn}

\author{Zhihuan Song}
\ead{songzhihuan@zju.edu.cn}

\address{State Key Laboratory of Industrial Control Technology, College of Control Science and Engineering, Zhejiang University, Hangzhou, 310027, P.~R.~China}

\cortext[cor1]{Corresponding author Xinmin Zhang}

\begin{abstract}
  With the development of intelligent manufacturing and the increasing complexity of industrial production, root cause diagnosis has gradually become an important research direction in the field of industrial fault diagnosis. However, existing research methods struggle to effectively combine domain knowledge and industrial data, failing to provide accurate, online, and reliable root cause diagnosis results for industrial processes. To address these issues, a novel fault root cause diagnosis framework based on knowledge graph and industrial data, called Root-KGD, is proposed. Root-KGD uses the knowledge graph to represent domain knowledge and employs data-driven modeling to extract fault features from industrial data. It then combines the knowledge graph and data features to perform knowledge graph reasoning for root cause identification. The performance of the proposed method is validated using two industrial process cases, Tennessee Eastman Process (TEP) and Multiphase Flow Facility (MFF). Compared to existing methods, Root-KGD not only gives more accurate root cause variable diagnosis results but also provides interpretable fault-related information by locating faults to corresponding physical entities in knowledge graph (such as devices and streams). In addition, combined with its lightweight nature, Root-KGD is more effective in online industrial applications.
\end{abstract}


\end{frontmatter}



\section{Introduction}

To ensure the stable operation of industrial equipment, fault diagnosis has been widely applied to industrial production. In the past few decades, numerous fault diagnosis studies have focused on fault detection and classification, which have been successfully implemented in actual industrial production\cite{ge2013review, zhou2020data, jiang2017semi}. However, with the evolution of modern industrial technology towards greater scales and increased complexity, the equipment and interconnections involved in the industrial processes become more complicated, with a larger number of variables that exhibit complex coupling relationships. Conventional fault diagnosis methods struggle to achieve effective fault localization, which makes it difficult to implement timely and effective corresponding measures. Therefore, in recent years, the root cause diagnosis of faults has gradually become a significant research topic.

The contribution-based fault diagnosis methods are commonly utilized for the identification of key variables. Although variables with high fault contributions are not necessarily the root causes, the identification of key variables can significantly narrow down the scope of potential root causes. The traditional contribution plot is developed based on Principal Component Analysis (PCA) \cite{abdi2010principal}, which calculates the contribution score for each variable by decomposing the fault detection indicators, like $\mathrm{T^2}$, $\mathrm{SPE}$, as well as their combinations. Reconstruction-Based Contribution (RBC) is proposed to address the issue of the smearing effect in the contribution plot calculated by PCA \cite{alcala2009reconstruction}. Moreover, Qian et al. \cite{qian2020locally} employed a back-propagation (BP) algorithm to describe the propagation of fault information and applied Autoencoder (AE) to build a deep learning model for fault diagnosis to extract nonlinear features.

Currently, many algorithms in root cause diagnosis tasks are implemented based on causal inference. These methods analyze causal relationships between variables to identify root causes, which can be considered as a causal reasoning task for multivariate time series with industrial data. Transfer Entropy (TE) \cite{schreiber2000measuring} and Granger causality (GC) \cite{granger1969investigating} are classical methods for causal inference and serve as the research foundation for many methods in this field. Symbolic Transfer Entropy (STE) \cite{staniek2008symbolic} excels at capturing fluctuations in time series data and deals with nonlinear and non-stationary issues. Neural Granger Causality (NGC) \cite{tank2021neural} applies structured multilayer perceptron (MLP) and recurrent neural network (RNN) to achieve the effective capture of long-range dependencies between series. Neural Graphical Modeling (NGM) \cite{bellot2021neural} is an enhanced version of NGC, which can realize accurate inference under sample irregularity and nonlinear problems.

In the field of root cause diagnosis, numerous studies have been conducted to adaptively adjust causal inference methods. For example, Bauer et al. \cite{bauer2006finding} applied TE to fault propagation path analysis in the industrial field and designed a method for automatically generating causal maps of process variables. Chen et al. \cite{chen2018root} applied the multivariate GC technique to construct the causal map between process variables and designed the maximum spanning tree to identify the root cause. Lindner et al. \cite{lindner2019systematic} studied the guidelines for selecting the optimal TE parameters, and provided a robust program to accurately identify the propagation path of oscillations. Song et al. \cite{song2023mpge} used multi-layer convolutional neural networks (CNN) for nonlinear feature extraction of industrial data and characterized direct and indirect Granger causalities with multi-level predictive relationships to identify the root cause. It can be seen from the aforementioned works that most of the causal inference-based root cause diagnosis methods require extensive time series data for effective causal analysis, which is more suitable for off-line fault analysis, but can hardly provide real-time diagnosis results when performed in the online industrial application.

The fault diagnosis methods based on contribution and causal inference mainly depend on data-driven modeling. However, the lack of domain knowledge often leads to a deficiency in both accuracy and interpretability. Dynamic Bayesian Network (DBN) \cite{murphy2002dynamic} has been the main method for root cause diagnosis driven by data and knowledge in recent years. Yu et al. \cite{yu2013novel} proposed the networked process monitoring framework based on DBN for fault detection and root cause diagnosis. Zhang \cite{zhang2015dynamic} designed a dynamic uncertain causality graph based on DBN for knowledge representation and probabilistic reasoning to implement interpretable causal reasoning. Furthermore, dividing complex industrial processes into blocks is a widely adopted technique for integrating domain knowledge in root cause analysis. Dong et al. \cite{dong2023hierarchical} constructed a hierarchical causal graph based on TE and divided the process into several subblocks to reduce the causal connections, which can improve the effectiveness of root cause diagnosis. He et al. \cite{he2023causal} proposed a causal topology-based variable-wise generative model, which reduced the complexity of the model by dividing the variables into different groups to realize the device-level fault tracing.

Knowledge Graph (KG) is a method of structured representation of knowledge, composed of triples that can be expressed as (head entity, relation, tail entity) \cite{rossi2021knowledge}. Knowledge graph reasoning refers to the process of inferring unknown triples, predicated on the existing corpus of knowledge \cite{chen2020review}. The classical methods for knowledge graph reasoning include TransE \cite{bordes2013translating}, ComplEx \cite{trouillon2016complex}, RotateE \cite{sun2019rotate}, and so on. These methods typically focus on the static representation and reasoning of entities and relations, which are unable to adapt to the variety of fault modes in industrial processes. To apply knowledge graphs to the field of fault diagnosis, Han et al. \cite{han2023leveraging} constructed the knowledge graph based on expert knowledge of hot rolling line and history logs of fault maintenance, and they implemented a multi-hop Question Answering (Q\&A) System for fault diagnosis using reinforcement learning methods. Chi et al. \cite{chi2021distributed} designed a distributed knowledge inference framework for fault diagnosis in the industrial Internet of Things system, and applied the knowledge graph reasoning algorithm to the Tennessee Eastman Process \cite{downs1993plant}. Zhou et al. \cite{zhou2024causalkgpt} developed a causal quality-related knowledge graph (CQKG) in the field of aerospace product manufacturing, and enhanced a large language model with the CQKG using instruction fine-tuning. These knowledge graph fault diagnosis methods that rely on expert knowledge or fault logs can only perform reasoning or Q\&A based on inherent knowledge, and are unable to integrate with the characteristics of various fault samples of industrial data for online fault diagnosis. Ren et al. \cite{ren2022association} established the multi-level knowledge graph to achieve association hierarchical knowledge representation and utilized graph convolutional networks (GCN) to extract features of industrial data, enabling online fault monitoring of the plant-wide industrial process. Chen et al. \cite{chen2023tele} introduced KTeleBERT, a tele-knowledge enhanced re-training model in the field of communications. This model leverages expert knowledge graphs, log documents, and machine data for multi-task training of a pre-trained model, facilitating applications in various downstream tasks, including root cause analysis and fault chain tracking. Despite the effective encoding of multimodal data by the pre-trained model, the substantial data and computational resources they demand often hinder their application in industry. Furthermore, the current fault diagnosis methods that integrate knowledge graph and industrial data usually require deep learning models and lack interpretability.

To address the above issues, a novel framework for Root cause diagnosis based on Knowledge Graph and industrial Data (Root-KGD) is proposed, which can effectively combine the valuable information from domain knowledge and practical industrial data. For domain expert knowledge, we construct a Prior Industrial Knowledge Graph (PIKG) to combine the physical and conceptual connections between entities. The contribution-based fault diagnosis method is employed to represent fault features in industrial data, which are linked to the corresponding entities of PIKG to achieve the knowledge graph representation in different fault modes. Considering that the root cause node can be represented as root cause relations with any other nodes in the knowledge graph, we design the score function for the root cause node and implement knowledge graph reasoning based on the Ripple Fault Propagation Algorithm (RFPA), ultimately obtaining the ranking results for each entity as the root cause. The main contributions of this paper are summarized as follows:

1) A novel framework named Root-KGD is proposed to implement knowledge- and data-driven fault root cause diagnosis with strong interpretability and lightweight performance, which is more applicable for online fault diagnosis.

2) We propose a knowledge graph root cause reasoning method that leverages the entity fault features. Compared to traditional knowledge graph reasoning methods, this method can integrate the fault features varying with the collected fault samples of industrial data.

3) RFPA is proposed to describe the root cause reasoning score of an entity as a root cause node for other entities, which is implemented through the structural features and the fault features of the knowledge graph with strong interpretability.

The rest of this paper is organized as follows. Section 2 reviews the classical contribution-based fault diagnosis method RBC. In Section 3, the proposed Root-KGD framework and RFPA are presented. In Section 4, two industrial processes, Tennessee Eastman Process (TEP) \cite{downs1993plant} and Multiphase Flow Facility (MFF) \cite{ruiz2015statistical}, are employed to verify the performance of the proposed method. Finally, the conclusions are made in Section 5.

\section{Revisit of Reconstruction-Based Contribution}\label{S2}
RBC \cite{alcala2009reconstruction} is mainly achieved by reconstructing the fault detection index along the variable direction. Compared to the traditional contribution plot, RBC can better suppress the smearing effect by fault reconstruction using the propagation of fault information throughout the model, resulting in more accurate diagnosis performance.

Specifically, for an industrial dataset denoted as $X\in \mathbb{R}^{m\times n}$, where $n$ is the number of variables, and $m$ represents the sample number, the sample covariance matrix $S$ can be calculated as:
\begin{equation}
    S=\frac1{m-1}X^TX
\end{equation}
where $X^T$ represents the transpose of the matrix X.

The covariance matrix can be transformed to the loading matrix of PCA by eigen decomposition, shown as follows\cite{abdi2010principal}:
\begin{equation}
    S=P\Lambda P^T+\tilde{P}\tilde{\Lambda}\tilde{P}^T
\end{equation}
where $\Lambda$ and $\tilde{\Lambda}$ denote the diagonal matrices for the principal and residual eigenvalues, while $P$ and $\tilde{P}$ represent the principal and residual loadings. 

Then, the sample vector $x$ can be decomposed as follows:
\begin{equation}
    x=\hat x+\tilde x=Cx+\tilde Cx
\end{equation}
where $C=PP^T$ and $\tilde C=\tilde P\tilde {P}^T$ indicate the projection matrices of principal component subspace and residual subspace.

The Hotelling’s $\mathrm{T^2}$ statistic and the squared prediction error ($\mathrm{SPE}$) are widely used as fault detection indicators. The RBC algorithm is proposed based on fault detection indicators and the reconstruction-based contribution of variable $x_i$ in $\mathrm{T^2}$ and $\mathrm{SPE}$ can be calculated as follows:
\begin{equation}
    \mathrm{RBC}_{i}^{\mathrm{T^2}}=x^{T} D \xi_{i} d_{i i}^{-1} \xi_{i}^{T} D x=\frac{\left(\xi_{i}^{T} D x\right)^{2}}{d_{i i}}
\end{equation}
\begin{equation}
    \label{equ:spe_rbc}
    \mathrm{RBC}_{i}^{\mathrm{SPE}}=x^{T}\tilde{C}\xi_{i}\big(\xi_{i}^{T}\tilde{C}\xi_{i}\big)^{-1}\xi_{i}^{T}\tilde{C}x=\frac{\big(\xi_{i}^{T}\tilde{C}x\big)^{2}}{\tilde{c}_{ii}}
\end{equation}
Here, $\xi_i$ symbolizes the direction of the fault. The  $D$ is a positive semidefinite matrix, defined as $D=P\Lambda^{-1} P^T$. $d_{ii}$ corresponds to the $i$-th diagonal element of $D$, while $\tilde{c}_{ii}$ is identified as the $i$-th diagonal element of $\tilde{C}$.

\section{Methodology}\label{S3}

\subsection{Framework of the proposed method}

\begin{figure}[htbp]
    \centering
    \includegraphics[width=1\linewidth]{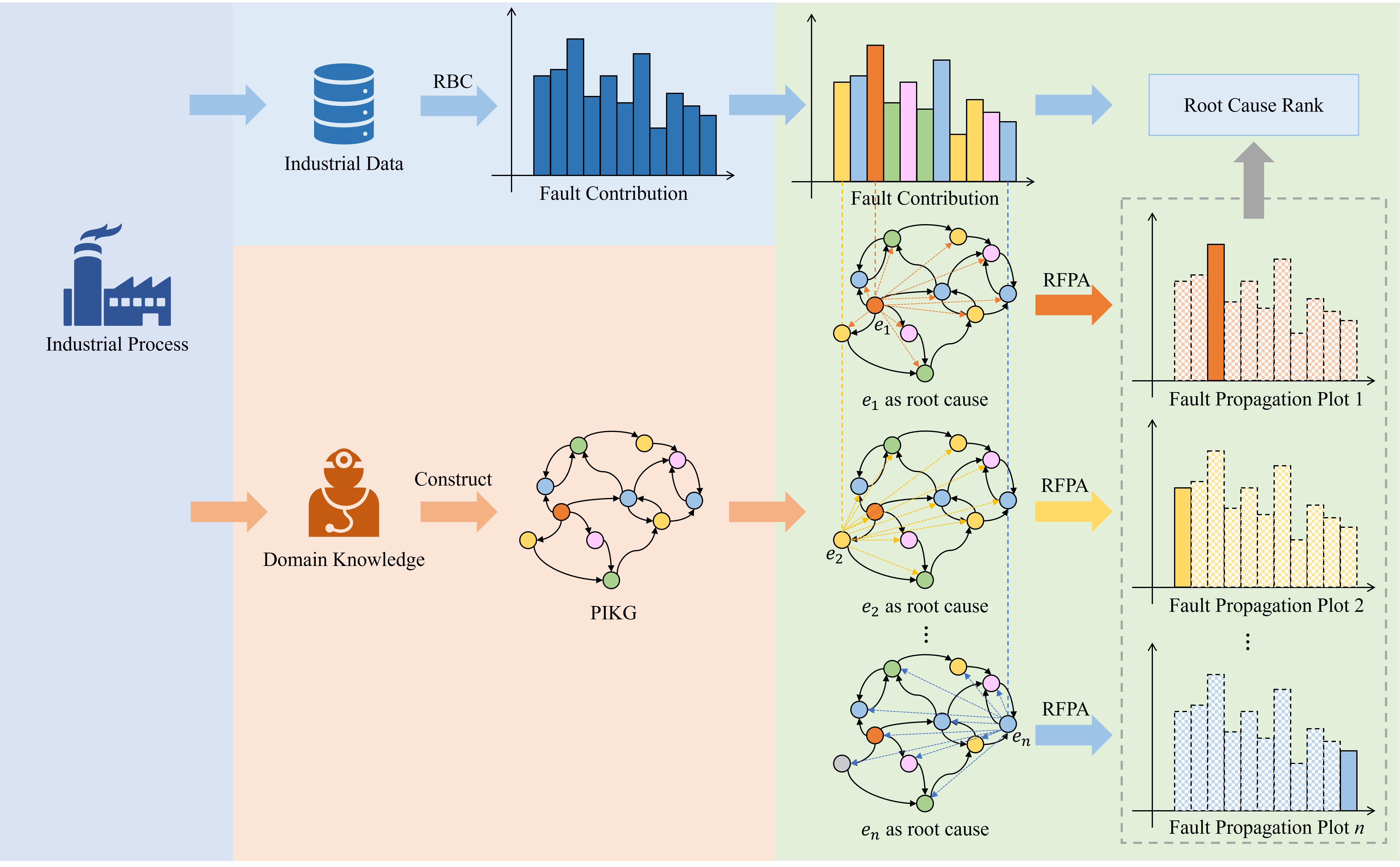}
    \caption{
        \label{fig:Root-KGD}  
        Framework of Root-KGD.}
\end{figure}

The proposed root cause diagnosis framework, Root-KGD, for industrial process faults is shown in Fig. \ref{fig:Root-KGD}, which is primarily composed of three steps.

\textbf{(1) Fault feature extraction of industrial data:}
Variable contributions can be used to represent the fault features varying with the collected samples from industrial data and reflect the impact of each variable on the operation of the system. The contribution score of each variable is commonly generated by data-driven contribution-based fault diagnosis methods. In this paper, the RBC algorithm is employed, and other alternative algorithms are also applicable.

\textbf{(2) The construction of PIKG:}
To encode the domain knowledge of the industrial process effectively, we develop a method for constructing PIKG, which connects entities such as devices and variables through physical or conceptual relations. PIKG describes the operational logic of the industrial system in the form of structured triples and constructs key connections for data and knowledge.

\textbf{(3) Root Cause Reasoning:}
To effectively represent industrial knowledge that varies with fault samples, we propose a knowledge graph reasoning method based on the entity fault features, and the fault contribution of each variable serves as the fault representation of the corresponding entity in PIKG. For the root cause node in a fault mode, it has a “root cause” relation with all other nodes in PIKG. The reasoning of the “root cause” relation can be described by the propagation of fault information from the source node to the target node. When a node has the highest score in the “root cause” relations with all other nodes, it indicates that the fault information propagated to these nodes is most closely aligned with the inherent fault information of the nodes themselves. As such, this node can be identified as the root cause node, as it is capable of reconstructing the fault features of the entire system through a combination of the structural features of the knowledge graph and its own fault attributes. Specifically, RFPA is designed to achieve the reasoning of the “root cause” relation. The root cause node rank is used to analyze the possibility of nodes as root causes.

\subsection{The construction of PIKG}

\begin{figure}[htbp]
    \centering
    \includegraphics[width=1\linewidth]{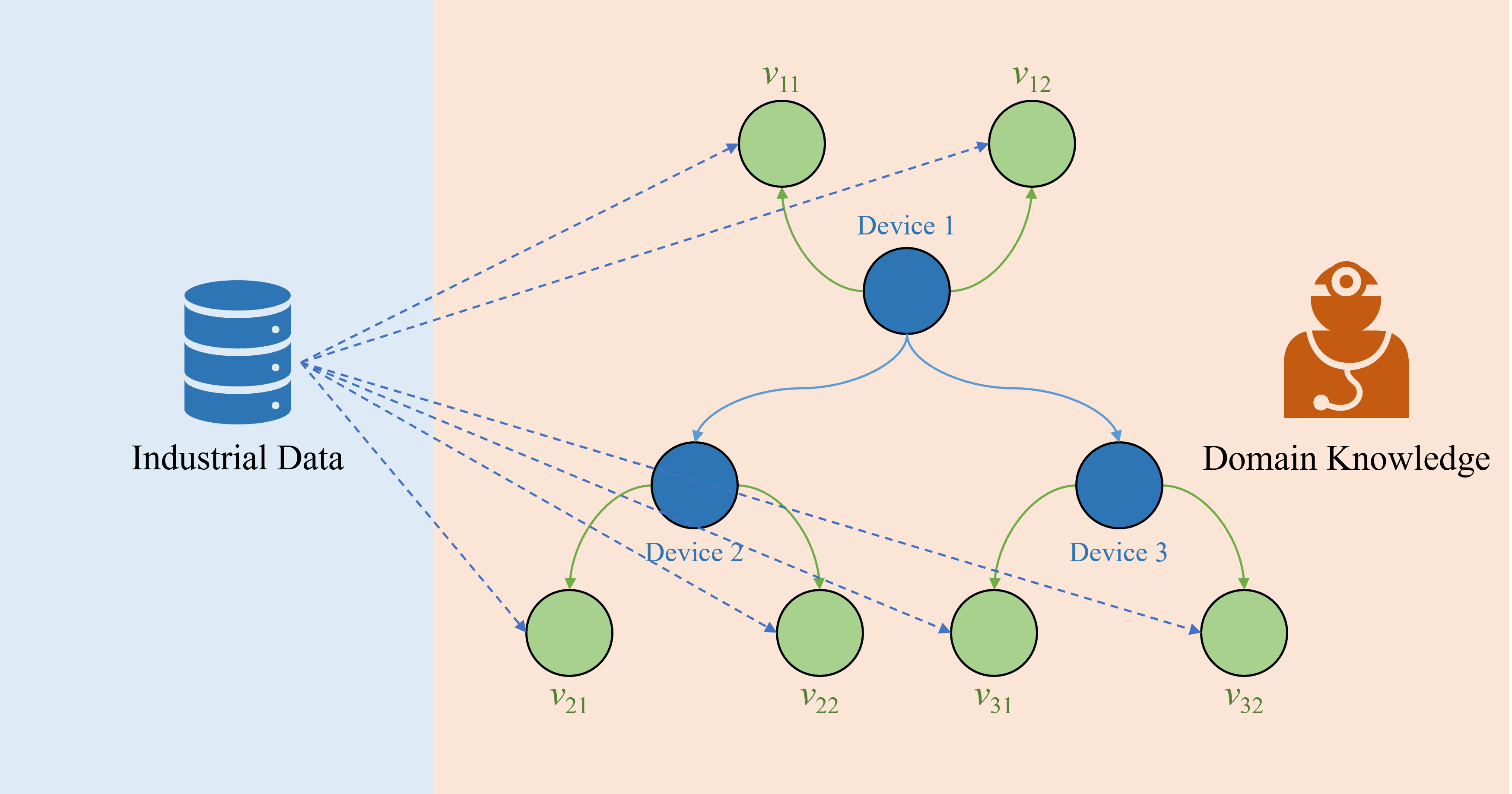}
    \caption{
        \label{fig:PIKG}  
        A simple example of defining entities and relations in PIKG.}
\end{figure}

In the domain of industrial fault diagnosis, constructing knowledge graphs based on a foundation of prior knowledge has become a common approach\cite{ren2022association, chen2023tele}. Following this methodology, we represent the industrial process in the form of knowledge graphs to extract the inherent features of the industrial system. Specifically, we introduce a method for constructing PIKG utilizing the existing domain expertise of the industry.

PIKG is represented as $G_p=(E,R,T)$, where $E$ is a set of entities (nodes in the graph), and $R$ is a set of relations (edges in the graph). $T$ denotes triples of $(e_h,r,e_t )$ that represents the connection of head entity ($e_h$), tail entity ($e_t$) and their relation ($r$), where $e_h,e_t\in E$, $r\in R$.

PIKG is composed of two entity types: physical entities and data entities, denoted as $E_p$ and $E_d$. $E_p$ represents the actual components within industrial processes, including devices, streams, and materials. $E_d$ corresponds to variables in the industrial data and indicates the operational status of physical entities, which serves as the critical connector between domain knowledge and industrial data. In other words, the industrial data associated with $E_d$ can be regarded as the temporal attributes of the entity.

The relations in PIKG are established based on the interactions between the head and tail entities. There are various types of relations expressed as $R=\{r_1,r_2,...,r_{n_r}\}$.Typically, there are inherent connections between different physical entities, while data entities often have status representation relations with their corresponding physical entities.

As shown in Fig. \ref{fig:PIKG}, the example entities and relations within PIKG are formed by nine nodes and their associated relations. The relations among the three physical entities (Device 1, Device 2, and Device 3) illustrate their physical connections, and the data entities ($v_{11}$, $v_{12}$, $v_{21}$, $v_{22}$, $v_{31}$, and $v_{32}$) are connected to their respective devices, indicating their operational status.

\subsection{Ripple fault propagation algorithm based on knowledge graph}

\begin{figure}[htbp]
    \centering
    \includegraphics[width=1\linewidth]{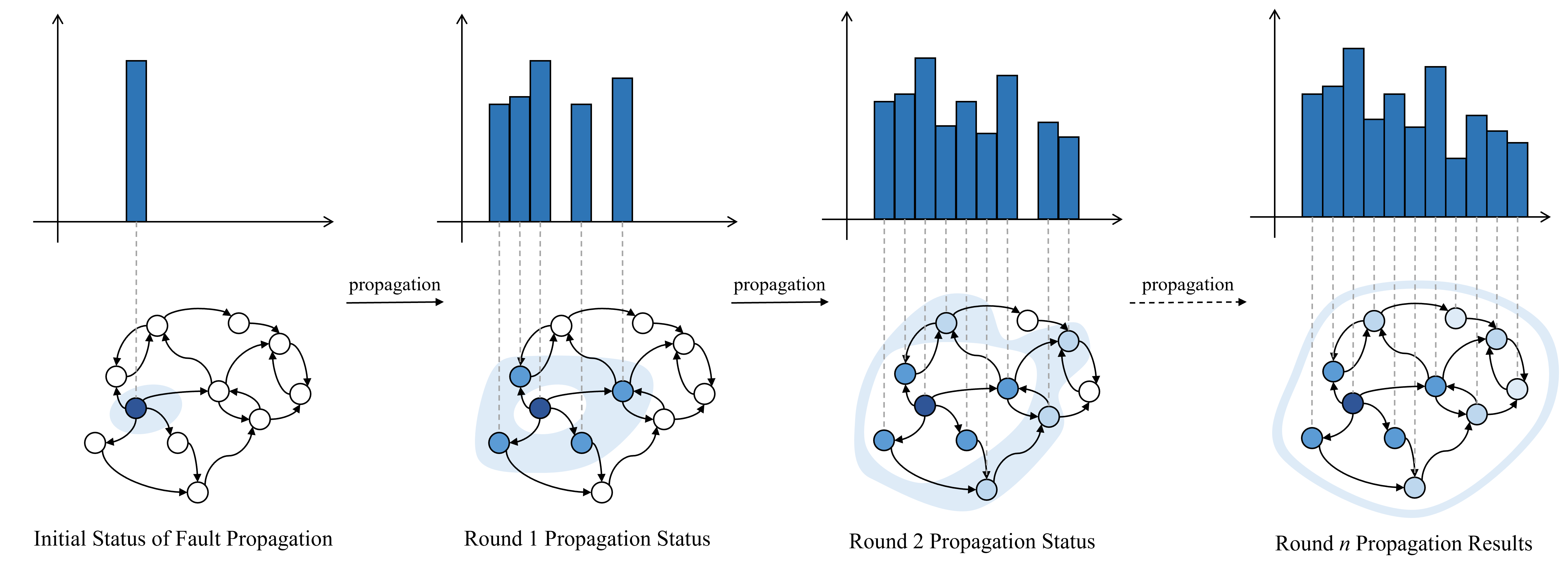}
    \caption{
        \label{fig:RFPA}  
        Ripple Fault Propagation Algorithm.}
\end{figure}

By leveraging the pre-constructed knowledge graph PIKG, we model the fault propagation processes to perform the reasoning of the “root cause”, which can also be viewed as the analysis of the impact of a fault originating from a particular node on the rest of the system's nodes. Specifically, RFPA is proposed based on the combination of structural features and entity fault features in knowledge graphs, as shown in Fig. \ref{fig:RFPA}.

RFPA firstly selects a particular node as the source for the fault propagation, assuming that this node is the root cause node of the current system fault, and the given initial node is represented as $e_0 \in E$. The initial value of the fault quantity $s_0$ is assigned to the initial node, which is then propagated to its connected child nodes through the relational pathways defined within the knowledge graph. Each node that receives the fault information from $e_0$ becomes a new source, further propagating the fault to its subsequent child nodes. After multiple propagations, every node in the system will accumulate a certain level of fault quantity. This fault quantity will gradually stabilize as the propagation attenuates. The algorithm mimics the behavior of ripples in a pond, where the initial disturbance spreads outward and generates waves upon contact with subsequent nodes, ultimately reaching a stable state.

Actually, the efficiency and attenuation of fault propagation vary due to different types of relations. The extent of fault propagation attenuation corresponding to the different relations can be expressed as $D_f=\{d_{r_1},d_{r_2},...,d_{r_{n_r}} \}$, and the priority order of fault propagation can be expressed as $O_p=\{o_{r_1},o_{r_2},...,o_{r_{n_r}} \}$. For the triple $(e_h,r_i,e_t )$, the propagation formula for the fault quantity from $e_h$ to $e_t$ is as follows:

\begin{equation}
    \Delta s_t= s_h\cdot L_{path}\cdot L_{receive}
\end{equation}
where $\Delta s_t$ means the fault quantity acquired of $e_t$ after this round of propagation process, and $s_h$ means the fault quantity of $e_h$ before this round of propagation. $L_{path}$ is donated as the attenuation related to relation types.
\begin{equation}
    L_{path}=\exp(-\sigma_r d_{r_i})
\end{equation}
where $\sigma_r$ is a hyperparameter, and $L_{path}\in(0,1)$. 

$L_{receive}$ is represented as the attenuation related to the count of times that a node receives fault propagation, which avoids the explosion problem of node fault quantity during propagation.
\begin{equation}
    L_{receive}=\frac{1}{N_r[e_h]} 
\end{equation}
where $N_r[\cdot]$ is defined as the count of fault propagations each entity has received. 

Considering the varying efficiency of propagation between different relations, we set a priority order for each node to satisfy their sequential relationship during the propagation process as follows
\begin{equation}
   C_p^t =  C_p^h + o_{r_i}
\end{equation}
where $C_p^h$ represents the priority order of node $e_h$ to propagate faults, which is actually equal to the priority of the current round. $C_p^t$ means the priority order of node $e_t$ to propagate faults, implying that $e_t$ can only act as a source of fault propagation after $o_{r_i}$ rounds have passed. 

Furthermore, the termination criteria of RFPA are defined. For the triple $(e_h,r_i,e_t )$, the conditions for stopping fault propagation are as follows:
\begin{equation}
    \begin{cases}
        N_s [e_t ]  \le P_{max}
        \\\Delta s_t<\Delta s_{min}
        \end{cases}
 \end{equation}
where $N_s[\cdot]$ denotes the count of fault propagations initiated by each entity, $P_{max}$ refers to the upper limit of propagation times for the nodes, and $\Delta s_{min}$ is the threshold for the minimum quantity of fault propagation.

\begin{algorithm}[htbp]
    \SetAlgoLined 
	\caption{Ripple Fault Propagation Algorithm}
        \label{alg1}
	\KwIn{Given initial entity, $e_0$. The PIKG, $G_p=(E,R,T)$.}
	\KwOut{the set of fault propagation results for all entities, $S_{RFPA}[\cdot]$.}
	Define a priority queue $PQ$, with two attributes: entity and priority order. 
    
    Define the current count of priority order instances $C_p$. 
 
    Initialize $PQ$ to an empty queue and set $C_p$ to 0.

    \For {$e_i \in E$} {
        $N_s [e_i ]\gets 0$, $N_r [e_i ]\gets 0$, $S_{RFPA} [e_i ]\gets s_0$;
    }

    Push ($PQ$, [\rm {entity}: $e_0$, priority order: $C_p$] ), $N_s [e_0]\gets N_s [e_0 ]+1$.

	\While{$PQ$ is not empty}{
 	      $n_{PQ}$  $\gets$ the number of elements in $PQ$;
        
		\For {$i\gets 0$ \rm{\textbf{to}} $n_{PQ}$}{
                $(e_v  ,C_v) \gets$ Pop ($PQ$), $N_s [e_v] \gets N_s [e_v ]+1$;

                \For {$(e_v,r_c,e_c)\in T $}{
                    \If{$N_s [e_v ]  \le P_{max}$}{

                        $\Delta s_c= S_{RFPA} [e_v]\cdot L_{path}\cdot L_{receive}$;

                        \rm{\textbf{if}} $\Delta s_c<\Delta s_{min}$ \rm{\textbf{then}} \rm{\textbf{break}};
                        
                        $S_{RFPA} [e_c] \gets S_{RFPA} [e_c ]+ \Delta s_c$, $N_r [e_v] \gets N_r [e_v ]+1$; 

                        Push ($PQ$, [entity: $e_c$, priority order: $C_p+ o_{r_c}$] );
                    }
                }
                $	C_p \gets C_p+1$;
            }
	}
\end{algorithm}

The implementation of RFPA leverages a priority queue to manage the fault propagation sequence of entities, which is summarized in Algorithm \ref{alg1}. For an initial entity $e_k \in E$, the set of fault propagation results for all entities $S_{RFPA}[\cdot]$ can be transformed into a sequence representation, denoated as $S_{RFPA}^k$.

\subsection{Root cause reasoning}

For entity $e_A$ in PIKG, it can be assumed that $e_A$ is the root cause of the current fault mode. Therefore, for any entity $e_k$ in PIKG, and $e_k$ is not the root cause $e_A$, there exists the triple $(e_A,r_{root},e_k )$. $r_{root}$ in the triple indicates the “root cause” relation, and its relation scoring function can be denoted as:
\begin{equation}
    \label{equ:root_score}
    E\,(e_A,e_k )=L\,(\,rep\,(e_A ),rep\,(e_k )\,)=L\,(\,f_{root} \,(\,rep\,(e_A )\,),rep\,(e_k )\,)
\end{equation}
where $rep\,(\cdot)$ indicates the entity representation, which describes the fault information of the entity. It can be represented by the fault contribution calculated by the data-driven contribution-based fault diagnosis model. $f_{root}\, (\cdot)$ indicates the reasoning score function of relation $r_{root}$, which predicts the representation of the tail entity based on the given representation of the head entity and the relation. It can be represented by the fault propagation results calculated by the RFPA algorithm from the source node $e_A$ to $e_k$. $L\,(\cdot)$ indicates the distance between the representations of two entities. The smaller the distance, the more likely $e_A$ is to be the root cause.

For all entities in the entire knowledge graph, the score for $e_A$ to be the root cause node can be calculated as follows:
\begin{equation}
    \mathrm{RootScore}(e_A)=\sum_{e_k\in E_v,e_k\neq e_A}{L\,(\,f_{root}\,(\,cont_A\,),cont_k\,)}
\end{equation}
where $cont_A$ and $cont_k$ represent the fault contribution of $e_A$ and $e_k$, respectively. 

Actually, not every entity has a corresponding fault contribution, therefore, in the selection of $e_k$, only the entities corresponding to the variables involved in the industrial dataset can be considered, which can be represented as $E_v=\{e_1,e_2,...,e_{n_v} \}$, where $E_v \subseteq E_d$ and $n_v$ is the count of variables in industrial datasets. The variables are denoted as $V=\{v_1,v_2,...,v_{n_v } \}$ and each element $e_i$ in $E_v$ corresponds to a variable in $V$. To quantify the impact of faults across these variables, RBC is utilized to generate the variable contribution, represented as $Cont=\{cont_1,cont_2,...,cont_{n_v } \}$. Specifically, RBC is calculated based on $\mathrm{SPE}$ statistics in practical calculations, as shown in formula (\ref{equ:spe_rbc}).

To calculate the score of the root cause node according to formula (\ref{equ:root_score}), we actually obtained it through aligning the fault propagation simulation sequence with the sequence of variable contribution, which can also be understood as the alignment between the fault mode simulated by RFPA algorithm and the fault mode identified by the data-driven contribution-based fault diagnosis model.

To implement the RFPA algorithm with entity $e_A$ as the initial node, the initial value of the fault quantity is set as $s_0=cont_A$, and then, the corresponding fault simulation sequence $S_{RFPA}^A$ is calculated. The aligned fault simulation sequence corresponding to $E_v$, denoted as $S_V^A=\{s_1^A,s_2^A,...,s_{n_v}^A \}$, is extracted from $S_{RFPA}^A$. Considering that some nodes do not have corresponding fault contributions, cosine similarity is used as the distance function $L \, (\cdot)$, avoiding the influence of initial value $s_0$ on the results in the RFPA algorithm When dealing with the nodes without corresponding fault contributions, $s_0$ can be set as a constant in the RFPA algorithm. The score of the root cause node can be calculated as follows:
\begin{equation}
    \mathrm{RootScore}(e_A)=sim(S_V^A,Cont)=\frac{S_V^A\cdot Cont}{||S_V^A||\times||Cont||}
\end{equation}

It can be seen that the designed $\mathrm{RootScore}$ essentially evaluates the alignment extent between the fault mode induced by each initial node in the RFPA algorithm and the fault mode described by RBC. In addition, the results also indicate the possibility that $e_A$ is the root cause node. Therefore, by ranking the root cause scores of the nodes within PIKG, the node with the highest similarity is considered the most likely root cause of the current fault mode of the system.

\section{Case studies}\label{S4}

The proposed method is validated in two industrial datasets, the Tennessee Eastman Process (TEP) and the Multiphase Flow Facility (MFF), respectively. Moreover, a detailed implementation of the proposed Root-KGD framework is presented. First, PIKG was constructed according to the industrial processes, including multiple entity and relation types. To better describe relations between units, numbers of extra entities, such as stream 14, are also added in the established PIKG. In addition, the parameters designed for Root-KGD are mainly in two parts: RBC and RFPA. The fault contributions are derived from the average results of employing the RBC algorithm on the first 100 fault samples after the fault occurrence. Due to the difference in the scale of each RBC result, original contribution scores based on RBC are further normalized to the contribution rate. The parameter involved in RBC is the principal component ratio $r_{pc}$. Furthermore, the RFPA algorithm involves parameters $D_f$, $O_p$ and $\sigma_r$, which can collectively be denoted as $\theta=[d_{r_1},d_{r_2},...,d_{r_{n_r}};o_{r_1},o_{r_2},...,o_{r_{n_r}};\sigma_r]$. For comparison of experimental results, two classic causal reasoning methods, GC and TE, are used as comparison methods. The subsequent variables of causal reasoning are represented by the variables with higher fault contribution scores and key variables obtained from prior knowledge.

\subsection{Tennessee Eastman process (TEP)}

\begin{figure}[htbp]
    \centering
    \includegraphics[width=1\linewidth]{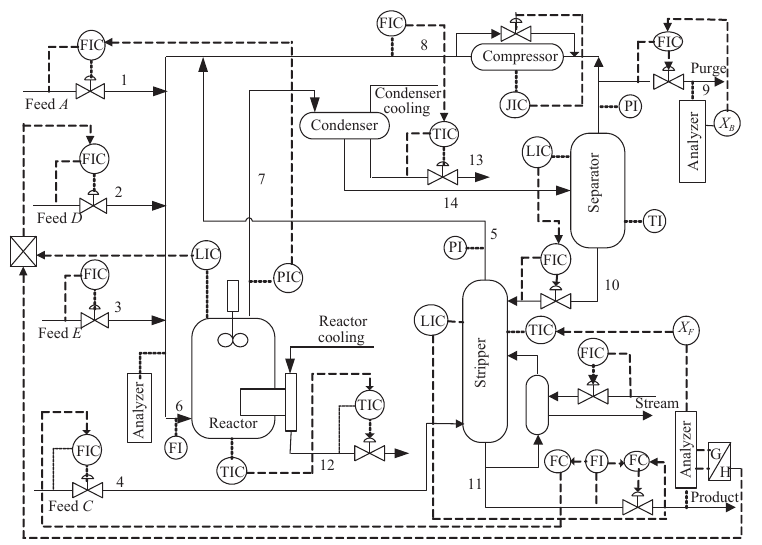}
    \caption{
        \label{fig:TEP}  
        Flowchart of the industrial TEP.}
\end{figure}

TEP \cite{downs1993plant} is a simulation system developed based on real chemical operations, and is a commonly used benchmark process in the process control domain. TEP includes 5 operating units: reactor, product stripper, vapor-liquid separator, product condenser, and recycle compressor. The system realizes the reaction processes among 7 different materials, involving 22 process measurement variables, 19 component measurement variables, and 12 process operating variables. The process flowchart of TEP is shown in Fig. \ref{fig:TEP}.

Based on the industrial TEP, we construct its corresponding PIKG. The number of entities and specified relation triples contained in PIKG is shown in Table \ref{tab:PIKG_TEP}.

\begin{table}[!htp]
    \centering
	\setlength{\abovecaptionskip}{0pt}%
	\setlength{\belowcaptionskip}{10pt}%
    \caption{Details of the PIKG constructed according to TEP.}
    \label{tab:PIKG_TEP}
    \resizebox{\columnwidth}{!}{
        \begin{tabular}{ccc}
            \toprule
            Concept          &Number       &Example\\
            \midrule
            Physical entity: Device	&5	&Reactor, Product Stripper\\
            Physical entity: Stream	&14	&Stream 1, Stream 6\\
            Physical entity: Substance	&7	&A, G\\
            Data entity: Variable	&51	&Reactor pressure ($x_7$), Reactor level ($x_8$)\\
            Triple with relation: State	&69	&(Reactor, State, Reactor pressure ($x_7$))\\
            Triple with relation: State of	&69	&(Reactor pressure ($x_7$), State of, Reactor)\\
            Triple with relation: Contain	&42	&(Stream 1, Contain, A)\\
            Triple with relation: Contained by	&42	&(A, Contained by, Stream 1)\\
            Triple with relation: Output	&44	&(Stream 6, Output, Reactor)\\
            Triple with relation: Generate	&7	&(A, Generate, G)\\
            \bottomrule
        \end{tabular}
    }

\end{table}

In the case study for the TEP, we calculate the fault contributions using the whole 52 variables ($v_1$-$v_{52}$). The parameters involved in the Root-KGD framework are shown in Table \ref{tab:Param_TEP}.

\begin{table}[htbp]
    \centering
	\setlength{\abovecaptionskip}{0pt}%
	\setlength{\belowcaptionskip}{10pt}%
    \caption{Parameters adopted by the study case of TEP.}
    \label{tab:Param_TEP}
    \resizebox{\columnwidth}{!}{
        \begin{tabular}{ccccccccccc}
            \toprule
            $r_{pc}$	&$\sigma_r$	&$d_{State}$	&$d_{Output}$	&$d_{Contain}$	&$d_{Generate}$	&$o_{State}$	&$o_{Output}$	&$o_{Contain}$	&$o_{Generate}$\\
            \midrule
            0.5	&0.1	&1	&3	&5	&20	&1	&5	&8	&20\\
            \bottomrule
        \end{tabular}
    }
\end{table}

\subsubsection{Feed ratio step change fault case IDV(1)}

In the fault case IDV(1), in stream 4, there is a step change observed in the A/C feed ratio, while the composition of B remains constant. According to the fault description, it is evident that the root cause of the fault is an abnormality in the feed rate of A and C in stream 4. Therefore, the root cause variable of IDV(1) is process measurement variable $x_4$ (A and C feed in stream 4) or process operating variable $x_{45}$ (A and C feed flow in stream 4).

\begin{figure}[htbp]
    \centering

        \subfigure[]{
        \includegraphics[width=0.9\linewidth]{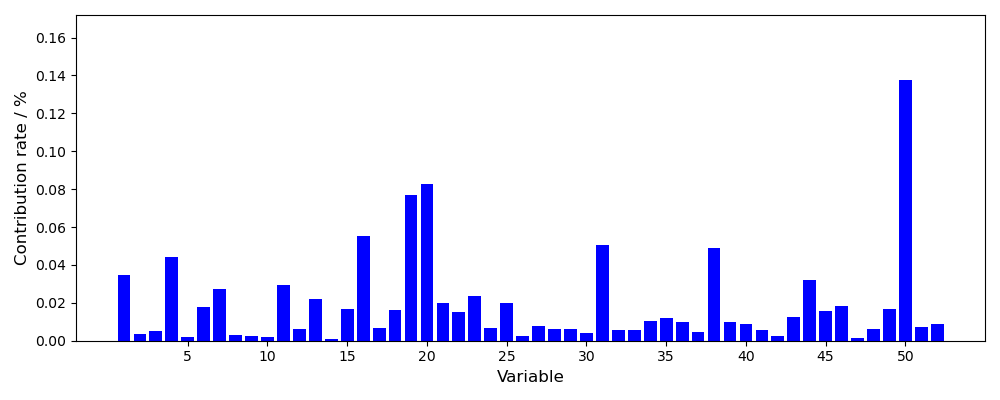} \label{fig:tep_d01_cont}
        }
        \subfigure[]{
        \includegraphics[width=0.45\linewidth]{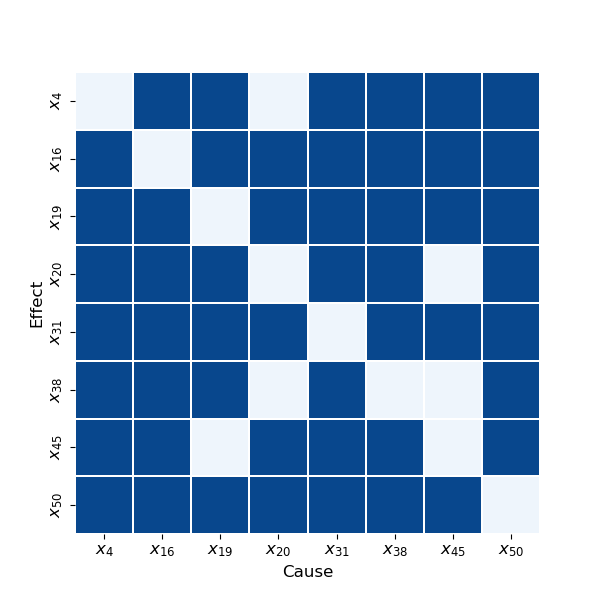} \label{fig:tep_d01_GC}
        }
        \subfigure[]{
        \includegraphics[width=0.45\linewidth]{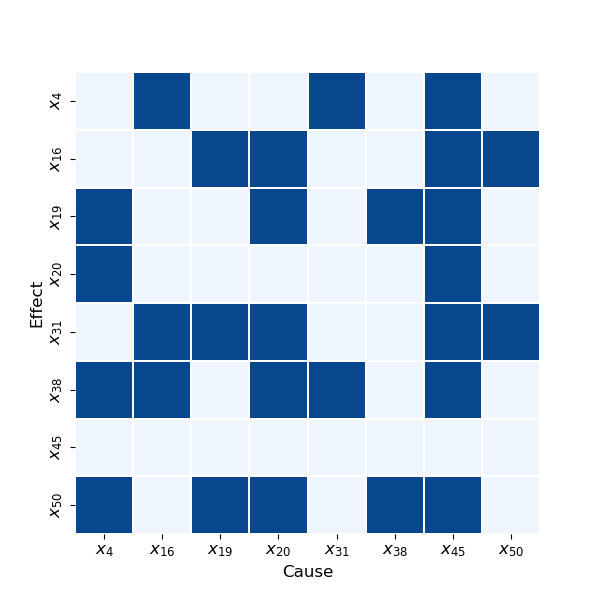} \label{fig:tep_d01_TE}
        }

    \caption{Root cause diagnosis results of different methods for fault case IDV(1): (a) RBC; (b) GC; (c) TE.}
    \label{fig:tep_d01}
\end{figure}

Fig. \ref{fig:tep_d01_cont} illustrates the fault diagnosis results of RBC. $x_{50}$ is identified as the key variable with the highest contribution, while neither $x_4$ nor $x_{45}$ achieved relatively high scores. Further, Fig. \ref{fig:tep_d01_GC}-\ref{fig:tep_d01_TE} shows the diagnostic performance of GC and TE. $x_4$ and $x_{45}$, as well as the other six variables with higher contributions from RBC, are selected as candidate variables. TE correctly calculated $x_{45}$ as the root cause, while $x_4$ is influenced by several other variables. In the test results of GC, there is a strong coupling relationship among the key variables, making it difficult to analyze the root cause variable.

\begin{table}[htbp]
    \centering
	\setlength{\abovecaptionskip}{0pt}%
	\setlength{\belowcaptionskip}{10pt}%
    \caption{Top10 nodes in the root cause score rank of Root-KGD for fault case IDV(1).}
    \label{tab:tep_d01}
    
    \begin{tabular}{cccc}
        \toprule
        Variable	&Score	&Stream and Device	&Score\\
        \midrule
        $x_{4}$	&0.55983	&Stream 4	&0.55582\\
        $x_{45}$	&0.55701	&Stream 1	&0.52006\\
        $x_{1}$	&0.52369	&Compressor	&0.50979\\
        $x_{44}$	&0.52341	&Stripper	&0.50052\\
        $x_{20}$	&0.51637	&Stream 14	&0.49190\\
        $x_{50}$	&0.51160	&Separator	&0.49190\\
        $x_{46}$	&0.50934	&Stream 10	&0.49124\\
        $x_{19}$	&0.50638	&Stream 6	&0.48299\\
        $x_{16}$	&0.50448	&Stream 8	&0.47883\\
        $x_{15}$	&0.50116	&Stream 5	&0.47825\\
        \bottomrule
    \end{tabular}

\end{table}

Table \ref{tab:tep_d01} tabulates root cause scores of the top 10 variables (data entities in PIKG) as well as device and stream entities by the Root-KGD framework. The variable $x_4$ and stream 4 obtain the highest scores, indicating the fault occurred on stream 4, and there is a problem with the feed of A and C, which is consistent with the actual fault mode. Furthermore, even though the variable $x_{45}$ has a very low contribution in the RBC results, its root cause score is only lower than $x_4$ in the Root-KGD framework, which also demonstrates the rationality of the proposed Root-KGD.

\subsubsection{Reactor variable step change fault case IDV(4)}

\begin{figure}[htbp]
    \centering

        \subfigure[]{
        \includegraphics[width=0.9\linewidth]{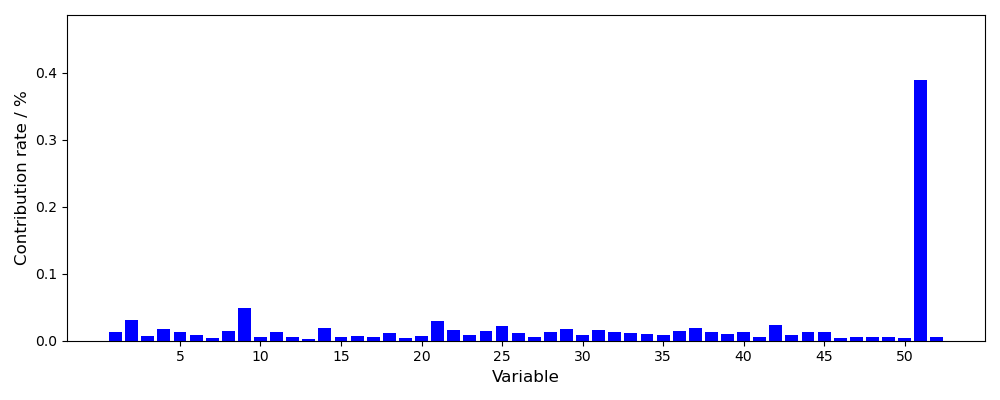} \label{fig:tep_d04_cont}
        }
        \subfigure[]{
        \includegraphics[width=0.45\linewidth]{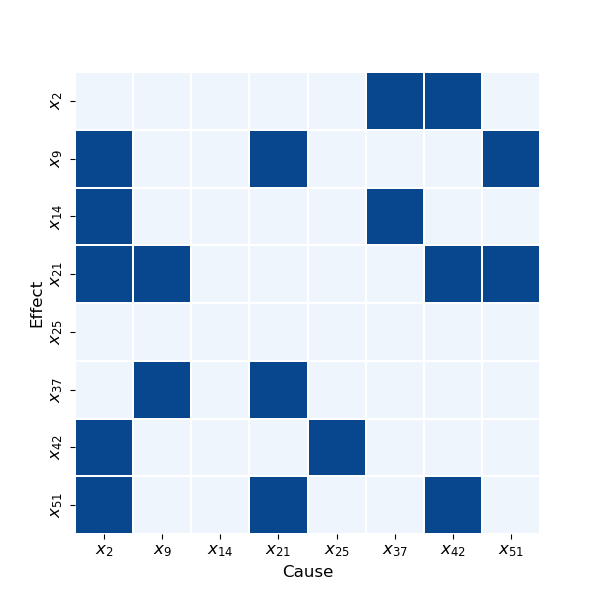} \label{fig:tep_d04_GC}
        }
        \subfigure[]{
        \includegraphics[width=0.45\linewidth]{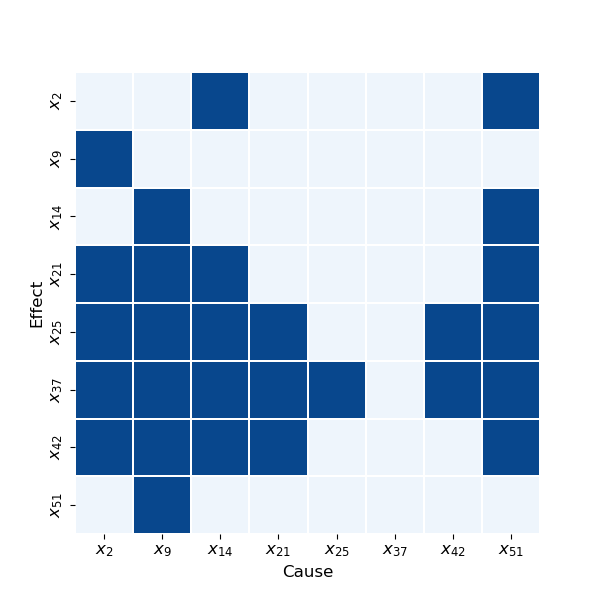} \label{fig:tep_d04_TE}
        }

    \caption{Root cause diagnosis results of different methods for fault case IDV(4): (a) RBC; (b) GC; (c) TE.}
    \label{fig:tep_d04}
\end{figure}

In the fault case IDV(4), the inlet temperature of the cooling water in the reactor has undergone a step change. However, there is no direct variable related to the inlet of the reactor cooling water. To ensure the normal operation of the reaction, the valve of stream 12 will adjust the cooling water flow rate of the reactor. Therefore, it can be observed that the root cause variable of the fault is the process operating variable $x_{51}$ (Reactor cooling water flow). 

Fig. \ref{fig:tep_d04_cont} shows the fault diagnosis results of RBC. $x_{51}$ is correctly identified as the key variable with the highest contribution. Further, Fig. \ref{fig:tep_d04_GC}-\ref{fig:tep_d04_TE} illustrates the diagnostic performance of GC and TE, and 8 variables with the highest contribution in RBC results are selected as candidate variables. GC mistakenly considered $x_{25}$ as the root cause. In the TE results, the identified $x_{51}$ and $x_9$ variables have coupling relationships with other variables.

\begin{table}[htbp]
    \centering
	\setlength{\abovecaptionskip}{0pt}%
	\setlength{\belowcaptionskip}{10pt}%
    \caption{Top10 nodes in the root cause score rank of Root-KGD for fault case IDV(4).}
    \label{tab:tep_d04}
    
    \begin{tabular}{cccc}
        \toprule
        Variable	&Score	&Stream and Device	&Score\\
        \midrule
            $x_{51}$	&0.55503	&Stream 12	&0.54986\\
            $x_{9}$	&0.54043	&Reactor	&0.54007\\
            $x_{8}$	&0.53837	&Stream 6	&0.43635\\
            $x_{7}$	&0.53769	&Stream 8	&0.41891\\
            $x_{21}$	&0.53414	&Stream 5	&0.41856\\
            $x_{6}$	&0.44509	&Stream 2	&0.39750\\
            $x_{26}$	&0.44455	&Stream 1	&0.38123\\
            $x_{25}$	&0.44317	&Stream 3	&0.38111\\
            $x_{23}$	&0.44214	&Compressor	&0.37694\\
            $x_{27}$	&0.44181	&Stream 11	&0.29466\\
        \bottomrule
    \end{tabular}

\end{table}

Table \ref{tab:tep_d04} shows root cause scores of the top 10 variables (data entities in PIKG) as well as device and stream entities by the Root-KGD framework. The variable $x_{51}$ and stream 12 obtain the highest scores, indicating the fault occurred on stream 12, and there is an issue with the cooling water flow rate, which can reconstruct the scene that the valve on stream 12 has been adjusted under feedback, which is consistent with the fault condition. In addition, the variable $x_9$ and reactor also obtain relatively high scores, which can illustrate the temperature inside the reactor is affected immediately due to changes in the cooling water temperature. In conclusion, the result of the proposed Root-KGD can better reconstruct the fault mode that cannot be directly located by variables.

\subsubsection{Feed step loss fault case IDV(6)}

\begin{figure}[htbp]
    \centering

        \subfigure[]{
        \includegraphics[width=0.9\linewidth]{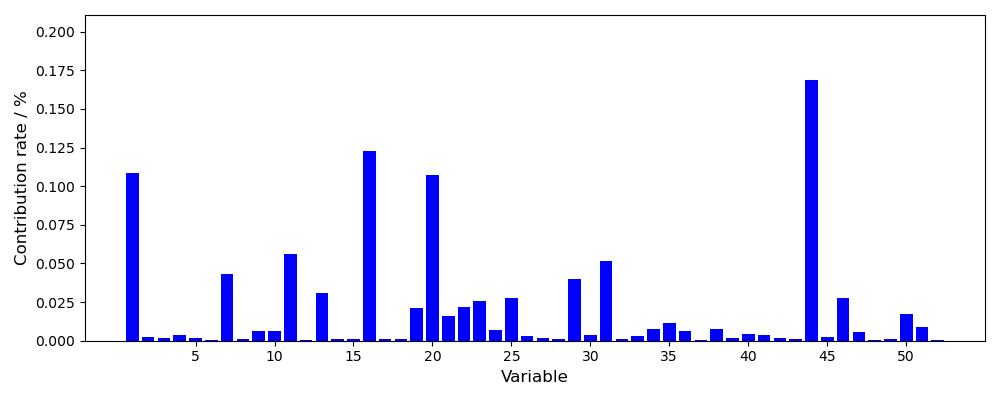} \label{fig:tep_d06_cont}
        }
        \subfigure[]{
        \includegraphics[width=0.45\linewidth]{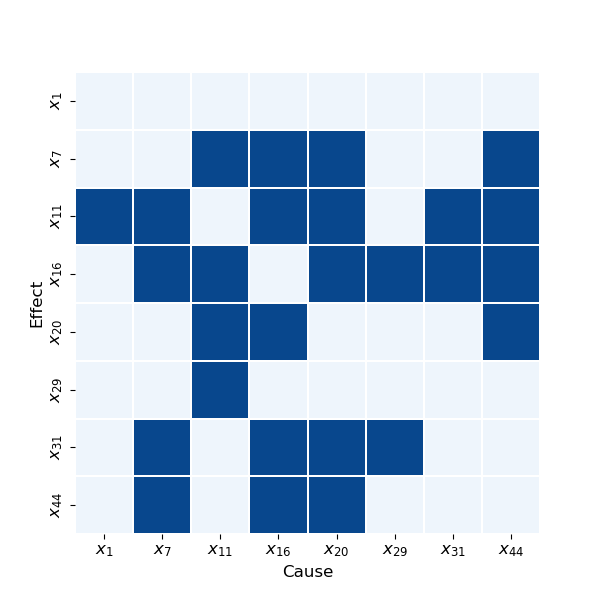} \label{fig:tep_d06_GC}
        }
        \subfigure[]{
        \includegraphics[width=0.45\linewidth]{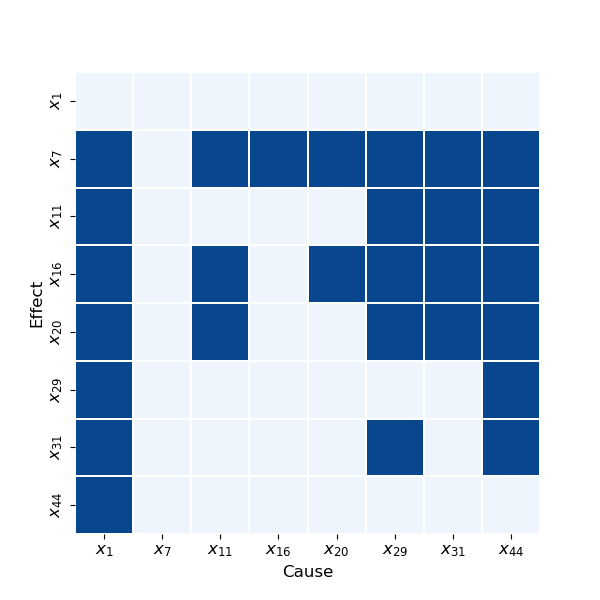} \label{fig:tep_d06_TE}
        }

    \caption{Root cause diagnosis results of different methods for fault case IDV(6) (a) The fault diagnosis results of RBC; (b) GC; (c) TE.}
    \label{fig:tep_d06}
\end{figure}

In the fault case IDV(6), in stream 1, the feed of A has a step loss. It is obvious that the root cause of the fault is the feed flow of A. Therefore, the root cause variable of IDV(6) is the process measurement variable $x_1$ (A feed in stream 1) or process operating variable $x_{44}$ (A feed flow in stream 1).

Fig. \ref{fig:tep_d06_cont} shows the fault diagnosis results of RBC. $x_{44}$ is correctly identified as the key variable with the highest contribution, while $x_1$, $x_{16}$ and $x_{20}$ have relatively high contributions. Fig. \ref{fig:tep_d06_GC}-\ref{fig:tep_d06_TE} further illustrates the diagnostic performance of GC and TE, and 8 variables with the highest contribution in RBC results are selected as candidate variables. Both GC and TE correctly identify $x_1$ as the root cause, and TE can identify $x_{44}$ as the root cause variable second only to $x_1$.

Table \ref{tab:tep_d06} tabulates root cause scores of the top 10 variables (data entities in PIKG) as well as device and stream entities by the Root-KGD framework. The variable $x_{44}$ and stream 1 obtain the highest scores, indicating the fault occurred on stream 1, and there is an issue with A feed flow, which is in line with the fault mode. In addition, the variable $x_1$ obtains a relatively high score second only to $x_{44}$, even though its contribution score in RBC is not so significant, which also proves the rationality of Root-KGD.

\begin{table}[htbp]
    \centering
	\setlength{\abovecaptionskip}{0pt}%
	\setlength{\belowcaptionskip}{10pt}%
    \caption{Top10 nodes in the root cause score rank of Root-KGD for fault case IDV(6).}
    \label{tab:tep_d06}
    
    \begin{tabular}{cccc}
        \toprule
        Variable	&Score	&Stream and Device	&Score\\
        \midrule            
            $x_{44}$	&0.50847	&Stream 1	&0.48986\\
            $x_{1}$	&0.50308	&Compressor	&0.41018\\
            $x_{20}$	&0.41767	&Stream 4	&0.38092\\
            $x_{46}$	&0.41082	&Stream 14	&0.37787\\
            $x_{4}$	&0.39648	&Separator	&0.37787\\
            $x_{45}$	&0.39639	&Stream 6	&0.37387\\
            $x_{23}$	&0.39187	&Stream 8	&0.36892\\
            $x_{11}$	&0.38024	&Stream 5	&0.36846\\
            $x_{13}$	&0.37886	&Condenser	&0.36032\\
            $x_{22}$	&0.37837	&Stream 7	&0.36032\\
        \bottomrule
    \end{tabular}

\end{table}

\subsubsection{Condenser variable random variation fault case IDV(12)}

In the fault case IDV(12), the inlet temperature of the cooling water in the condenser has encountered issues with random variables. However, there is no variable directly related to the condenser cooling water inlet. The impact of the change in the condenser cooling water temperature is most significant on the cooling effect of the corresponding flow stream, so the downstream flow stream 14 in the condenser will be directly affected, and then it will affect the temperature in the separator. Therefore, the variable closest to the root cause is $x_{11}$ (Product separator temperature).

\begin{figure}[htbp]
    \centering

        \subfigure[]{
        \includegraphics[width=0.9\linewidth]{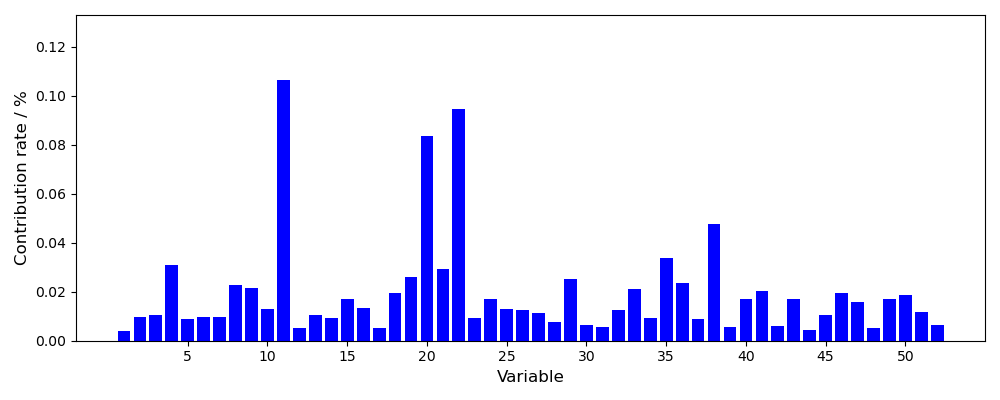} \label{fig:tep_d12_cont}
        }
        \subfigure[]{
        \includegraphics[width=0.45\linewidth]{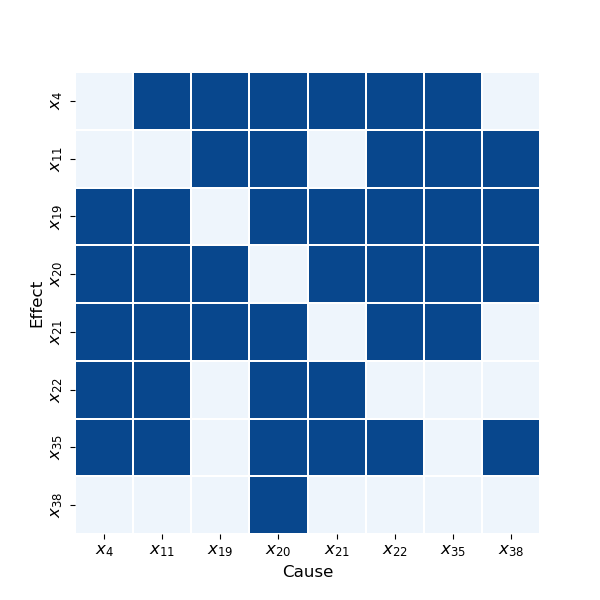} \label{fig:tep_d12_GC}
        }
        \subfigure[]{
        \includegraphics[width=0.45\linewidth]{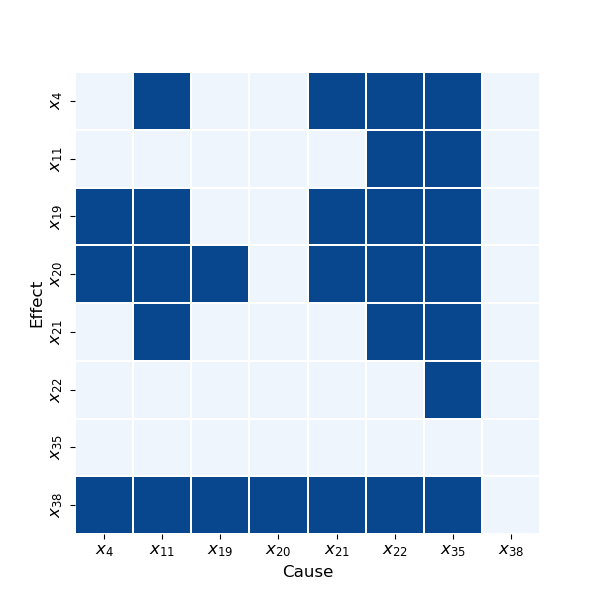} \label{fig:tep_d12_TE}
        }

    \caption{Root cause diagnosis results of different methods for fault case IDV(12): (a) RBC; (b) GC; (c) TE.}
    \label{fig:tep_d12}
\end{figure}

Fig. \ref{fig:tep_d12_cont} illustrates the fault diagnosis results of RBC. $x_{11}$ is correctly identified as the key variable with the greatest contribution. Fig. \ref{fig:tep_d12_GC}-\ref{fig:tep_d12_TE} further shows the diagnostic performance of GC and TE, and 8 variables with the highest contribution in RBC results are selected as candidate variables. TE incorrectly identified $x_{35}$ as the root cause variable, while in the results of GC, $x_{38}$ is identified as the root cause. These results are far from the actual fault mode.

\begin{table}[htbp]
    \centering
	\setlength{\abovecaptionskip}{0pt}%
	\setlength{\belowcaptionskip}{10pt}%
    \caption{Top10 nodes in the root cause score rank of Root-KGD for fault case IDV(12).}
    \label{tab:tep_d12}
    
    \begin{tabular}{cccc}
        \toprule
        Variable	&Score	&Stream and Device	&Score\\
        \midrule            
            $x_{11}$	&0.60934	&Stream 14	&0.60147\\
            $x_{22}$	&0.60843	&Separator	&0.60147\\
            $x_{13}$	&0.60187	&Stream 7	&0.58108\\
            $x_{12}$	&0.60147	&Condenser	&0.58108\\
            $x_{20}$	&0.58411	&Stream 13	&0.57967\\
            $x_{46}$	&0.57639	&Compressor	&0.57678\\
            $x_{52}$	&0.57623	&Reactor	&0.57261\\
            $x_{21}$	&0.57297	&Stream 12	&0.57003\\
            $x_{8}$	&0.57254	&Stream 6	&0.55105\\
            $x_{9}$	&0.57237	&Stream 8	&0.54255\\
        \bottomrule
    \end{tabular}

\end{table}

Table \ref{tab:tep_d12} shows root cause scores of the top 10 variables (data entities in PIKG) as well as device and stream entities by the Root-KGD framework. Variable $x_{11}$ achieved the highest root cause score, which is consistent with the fault mode. In addition, we can accurately locate the position where the fault occurred through the root cause scores on the stream and device entities. Stream 14 obtained the highest score, which reflects that there is a problem with the downstream flow stream of the condenser. In addition, the separator, stream 7 (the upstream flow stream of the condenser), the condenser itself, and stream 13 (the cooling water flow stream of the condenser) also obtained relatively high root cause scores. This can effectively locate the fault in the vicinity of the condenser. Combined with the temperature abnormality issues reflected in the root cause variables, it can effectively reconstruct the fault condition of the condenser cooling water abnormality. Compared to the traditional method of only locating the root cause variable, the approach of locating faults to devices and streams can more effectively analyze faults that are difficult to locate with variables.

\subsection{Multiphase Flow Facility (MFF)}

\begin{figure}[htbp]
    \centering
    \includegraphics[width=1\linewidth]{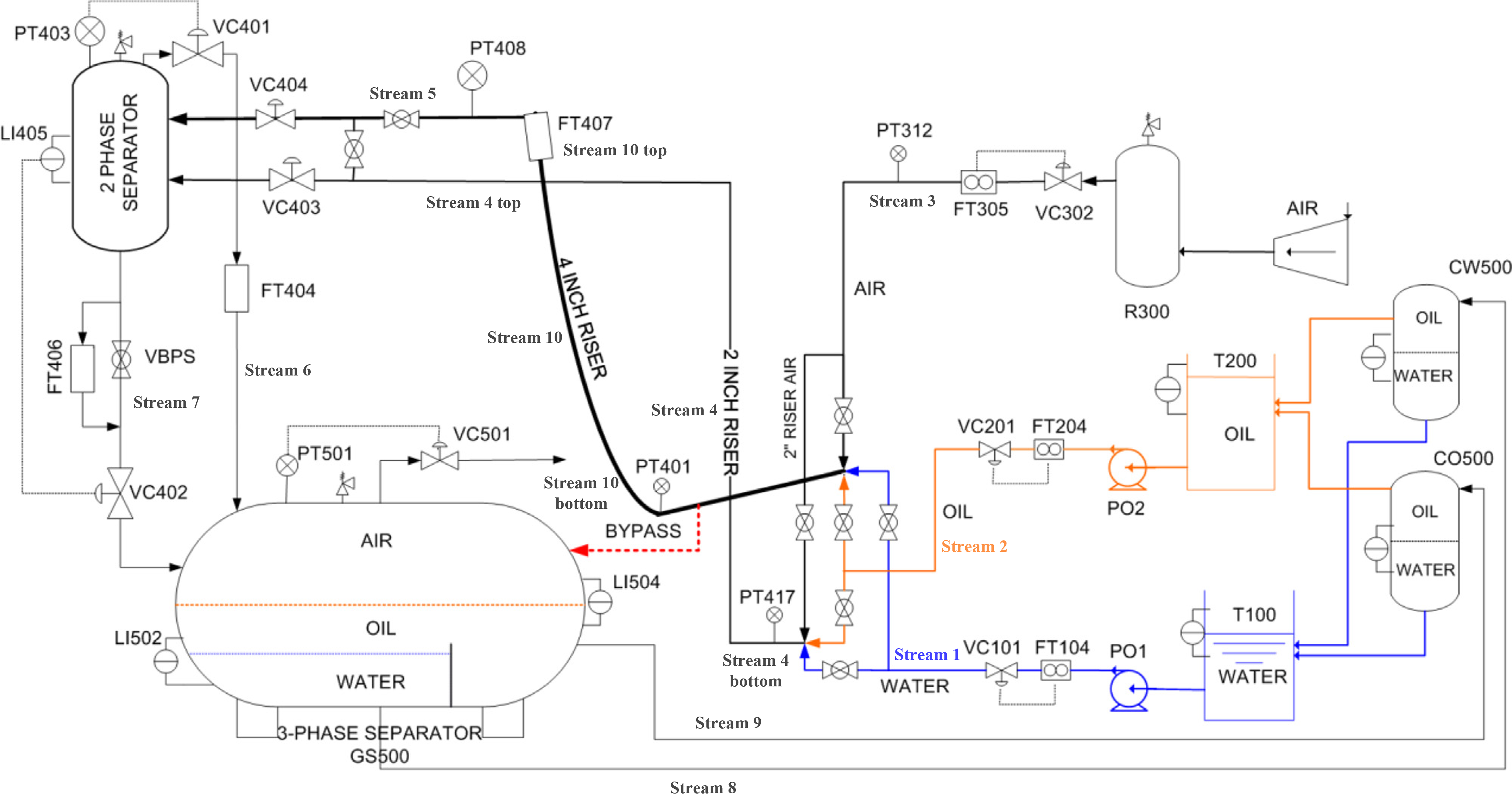}
    \caption{
        \label{fig:MFF}  
        Flowchart of the industrial process of Multiphase Flow Facility.}
\end{figure}

MFF \cite{ruiz2015statistical} is designed by Cranfield University and offers control and measurement of water, oil, and airflow rates within a pressurized environment. The MFF system includes a series of pipes with different diameters and configurations, a gas-liquid two-phase separator, and a three-phase separator on the ground, which includes 24 process variables. The MFF can be supplied with air, water, and oil at the desired rate in a single-phase or mixed phase, which can be mixed together and ultimately separated in the system. The process flow of MFF is shown in Fig. \ref{fig:MFF}.

Based on the industrial MFF process, we construct its corresponding PIKG. The number of entities and specified relation triples contained in PIKG is shown in Table \ref{tab:PIKG_MFF} .

\begin{table}[htbp]
    \centering
	\setlength{\abovecaptionskip}{0pt}%
	\setlength{\belowcaptionskip}{10pt}%
    \caption{Details of the PIKG constructed according to MFF. }
    \label{tab:PIKG_MFF}
    \resizebox{\columnwidth}{!}{
        \begin{tabular}{ccc}
            \toprule
            Concept          &Number       &Example\\
            \midrule
                Physical entity: Device	&16	&3 Phase Separator, T100\\
                Physical entity: Stream	&14	&Stream 1, Stream 8\\
                Physical entity: Substance	&3	&Water, Oil\\
                Data entity: Variable	&24	&Pressure in 3 Phase Separator ($x_5$), Density top riser ($x_{13}$)\\
                Triple with relation: State	&25	&(3 Phase Separator, State, Pressure in 3 Phase Separator ($x_5$))\\
                Triple with relation: State of	&25	&(Pressure in 3 Phase Separator ($x_5$), State of, 3 Phase Separator)\\
                Triple with relation: Contain	&40	&(T100, Contain, Water)\\
                Triple with relation: Contained by	&40	&(Water, Contained by, T100)\\
                Triple with relation: Output	&36	&(3 Phase Separator, Output, Stream 8)\\

            \bottomrule
        \end{tabular}
    }

\end{table}

In the case study for the MFF, $v_{24}$ (pressure in mixture zone 2-inch line, PT417) is only included in the analysis of fault case 6, while the first 23 variables are used in all the faulty cases studied \cite{ruiz2015statistical}.

The parameters involved in the Root-KGD framework adopted by the study case of MFF are shown in Table \ref{tab:Param_MFF}.

\begin{table}[htbp]
    \centering
	\setlength{\abovecaptionskip}{0pt}%
	\setlength{\belowcaptionskip}{10pt}%
    \caption{Parameters adopted by the study case of MFF.}
    \label{tab:Param_MFF}
    
    \begin{tabular}{ccccccccccc}
        \toprule
        $r_{pc}$	&$\sigma_r$	&$d_{State}$	&$d_{Output}$	&$d_{Contain}$	&$o_{State}$	&$o_{Output}$	&$o_{Contain}$\\
        \midrule
        0.8	&0.1	&1	&5	&10	&1	&1	&3\\
        \bottomrule
    \end{tabular}

\end{table}

\subsubsection{Top separator input blockage fault case 3}

In fault case 3, there is a blockage at the top separator inlet. The main device involved is the valve VC404, which controls the top separator inlet. Therefore, it can be considered that $x_7$ (Differential pressure over VC404) is the root cause variable of the fault.

\begin{figure}[htbp]
    \centering

        \subfigure[]{
        \includegraphics[width=0.9\linewidth]{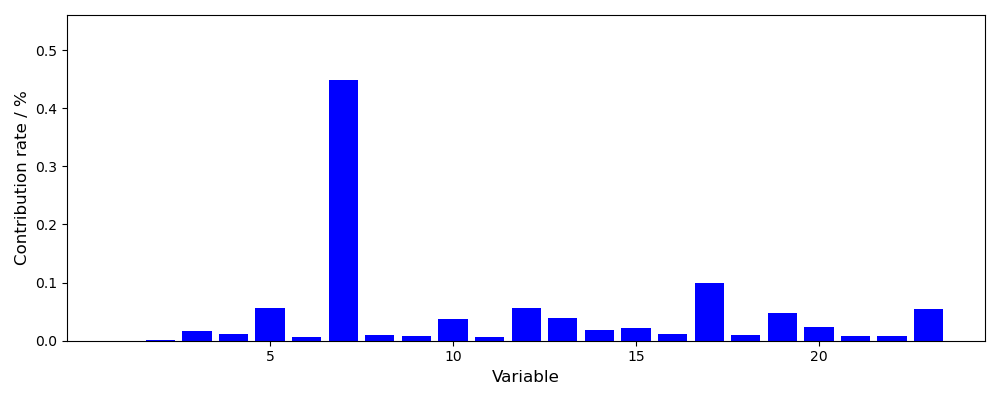} \label{fig:mff_s3_cont}
        }
        \subfigure[]{
        \includegraphics[width=0.45\linewidth]{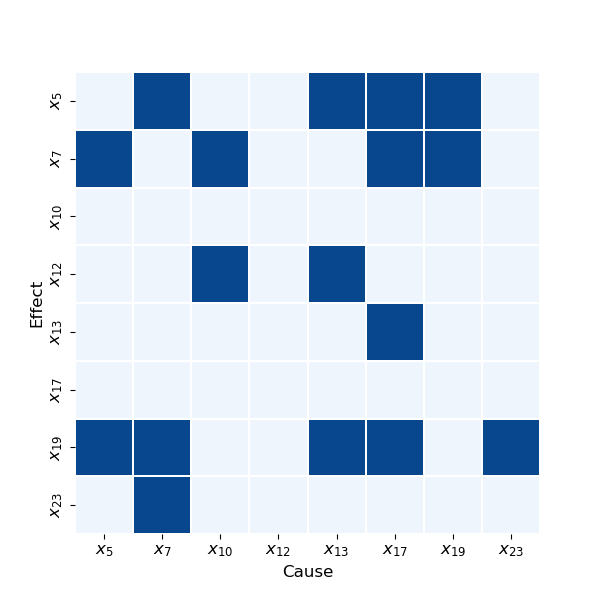} \label{fig:mff_s3_GC}
        }
        \subfigure[]{
        \includegraphics[width=0.45\linewidth]{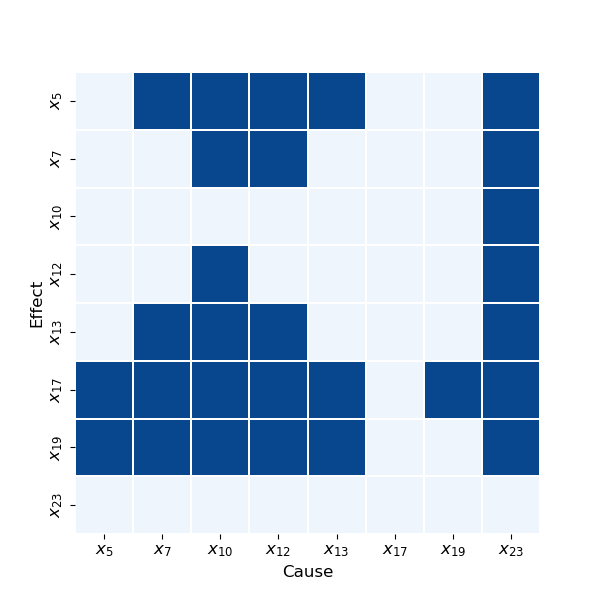} \label{fig:mff_s3_TE}
        }

    \caption{Root cause diagnosis results of different methods for fault case 3: (a) RBC; (b) GC; (c) TE.}
    \label{fig:mff_s3}
\end{figure}

Fig. \ref{fig:mff_s3_cont} illustrates the fault diagnosis results of RBC. The variable $x_7$ gets the highest contribution score, which is consistent with the fault condition. Fig. \ref{fig:mff_s3_GC}-\ref{fig:mff_s3_TE} further shows the diagnostic performance of GC and TE, and 8 variables with the highest contribution in RBC results are selected as candidate variables. GC identifies $x_{10}$ and $x_{17}$ as the root causes, while TE identifies $x_{23}$ as the root cause, none of which are consistent with the fault mode.

Table \ref{tab:mff_s3} tabulates root cause scores of the top 10 variables (data entities in PIKG) as well as device and stream entities by the Root-KGD framework. The variable $x_7$ is identified correctly. Stream 5 (the input stream of the top separator) and VC404 have the highest root cause scores, which can more accurately locate the fault compared to the positioning of key variables.

\begin{table}[htbp]
    \centering
	\setlength{\abovecaptionskip}{0pt}%
	\setlength{\belowcaptionskip}{10pt}%
    \caption{Top10 nodes in the root cause score rank of Root-KGD for fault case 3.}
    \label{tab:mff_s3}
    
    \begin{tabular}{cccc}
        \toprule
        Variable	&Score	&Stream and Device	&Score\\
        \midrule            
            $x_{7}$	&0.82881	&VC404	&0.81816\\
            $x_{4}$	&0.69096	&Stream 5	&0.81816\\
            $x_{11}$	&0.69003	&Stream 4 top	&0.73259\\
            $x_{17}$	&0.48531	&VC403	&0.73259\\
            $x_{12}$	&0.47611	&2 Phase Separator	&0.71345\\
            $x_{14}$	&0.46776	&Stream 9	&0.64359\\
            $x_{5}$	&0.46181	&Stream 8	&0.63460\\
            $x_{19}$	&0.45978	&Stream 4 bottom	&0.63223\\
            $x_{20}$	&0.45431	&Stream 4	&0.63223\\
            $x_{13}$	&0.45052	&CO500	&0.60896\\
        \bottomrule
    \end{tabular}

\end{table}

\subsubsection{Slugging condition fault case 5}

In fault case 5, the slugging occurred in the riser of the multiphase flow system, and the liquid accumulated at the bottom of the riser, obstructing the airflow and leading to a pressure difference in the riser. Therefore, the root cause should be located in the variable $x_6$ (Differential pressure between PT401 and PT408) of the riser (stream 10).

\begin{figure}[htbp]
    \centering

        \subfigure[]{
        \includegraphics[width=0.9\linewidth]{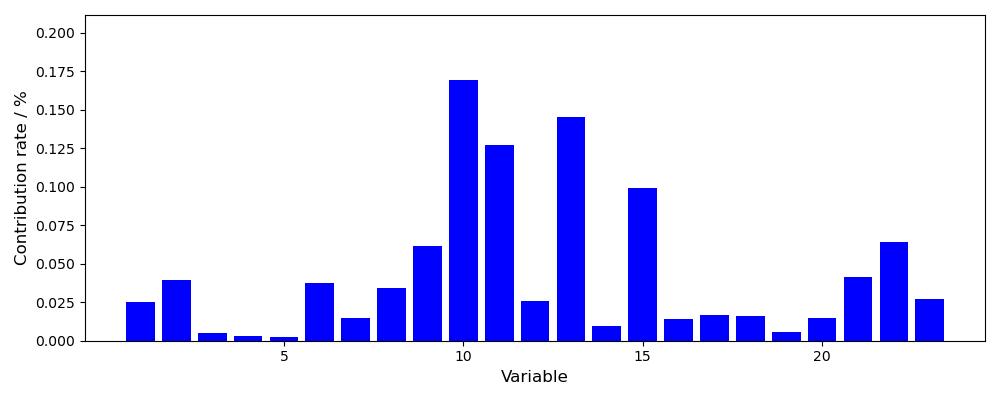} \label{fig:mff_s5_cont}
        }
        \subfigure[]{
        \includegraphics[width=0.45\linewidth]{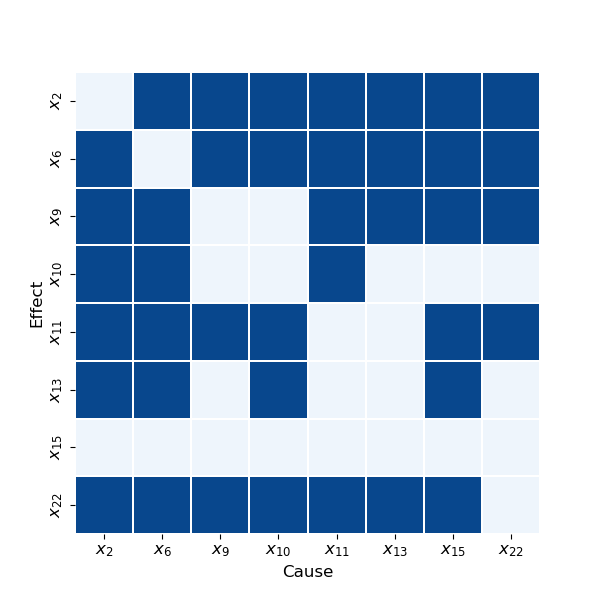} \label{fig:mff_s5_GC}
        }
        \subfigure[]{
        \includegraphics[width=0.45\linewidth]{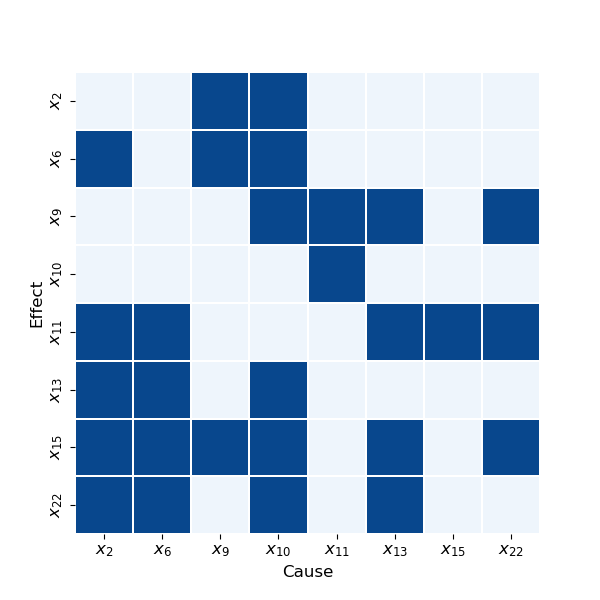} \label{fig:mff_s5_TE}
        }

    \caption{Root cause diagnosis results of different methods for fault case 5: (a) RBC; (b) GC; (c) TE.}
    \label{fig:mff_s5}
\end{figure}

Fig. \ref{fig:mff_s5_cont} shows the fault diagnosis results of RBC. $x_{10}$ (Flow rate top riser) is identified as the root cause, and $x_6$ achieves a relatively low contribution. Fig. \ref{fig:mff_s5_GC}-\ref{fig:mff_s5_TE} further shows the diagnostic performance of GC and TE, and 7 variables with the highest contribution in RBC results and the variable $x_6$ are selected as candidate variables. GC incorrectly identifies $x_{15}$ as the root cause variable, and TE does not effectively identify the root cause variable.

\begin{table}[htbp]
    \centering
	\setlength{\abovecaptionskip}{0pt}%
	\setlength{\belowcaptionskip}{10pt}%
    \caption{Top10 nodes in the root cause score rank of Root-KGD for fault case 5.}
    \label{tab:mff_s5}
    
    \begin{tabular}{cccc}
        \toprule
        Variable	&Score	&Stream and Device	&Score\\
        \midrule            
            $x_{6}$	&0.68013	&Stream 10	&0.68983\\
            $x_{2}$	&0.65874	&Stream 10 bottom	&0.67480\\
            $x_{10}$	&0.65228	&Stream 2	&0.66412\\
            $x_{13}$	&0.64772	&Stream 4	&0.65410\\
            $x_{22}$	&0.63106	&Stream 4 bottom	&0.65410\\
            $x_{15}$	&0.62845	&Stream 8	&0.63872\\
            $x_{16}$	&0.62233	&Stream 10 top	&0.63797\\
            $x_{3}$	&0.62055	&VC101	&0.63703\\
            $x_{9}$	&0.61874	&Stream 1	&0.62888\\
            $x_{18}$	&0.60683	&CO500	&0.62398\\
        \bottomrule
    \end{tabular}

\end{table}

Table \ref{tab:mff_s5} shows root cause scores of the top 10 variables (data entities in PIKG) as well as device and stream entities by the Root-KGD framework. The variable $x_6$ and the stream 10 are identified correctly, which indicates there is an issue with the pressure difference of the riser. Variable $x_2$ and the bottom section of stream 10 have relatively high root cause scores, which is consistent with the relationship between the slugging fault and the accumulation of water level at the bottom of the riser, reflecting the rationality of the proposed Root-KGD.

\subsubsection{Pressurization of the line fault case 6}

\begin{figure}[htbp]
    \centering

        \subfigure[]{
        \includegraphics[width=0.9\linewidth]{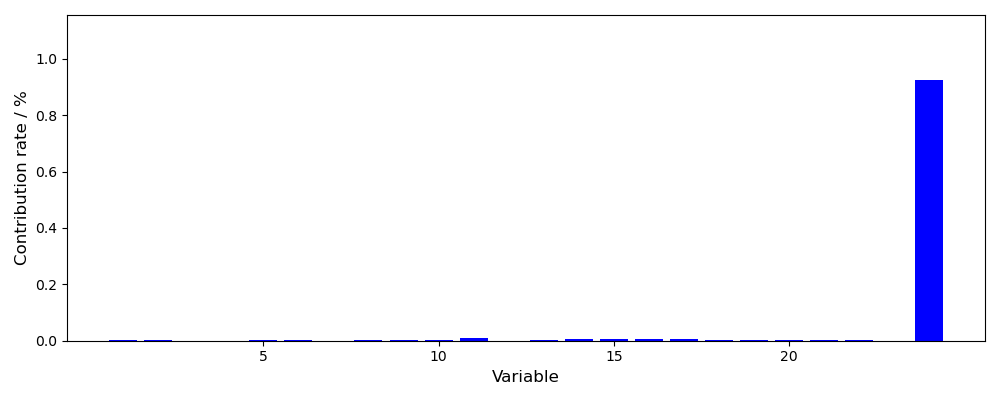} \label{fig:mff_s6_cont}
        }
        \subfigure[]{
        \includegraphics[width=0.45\linewidth]{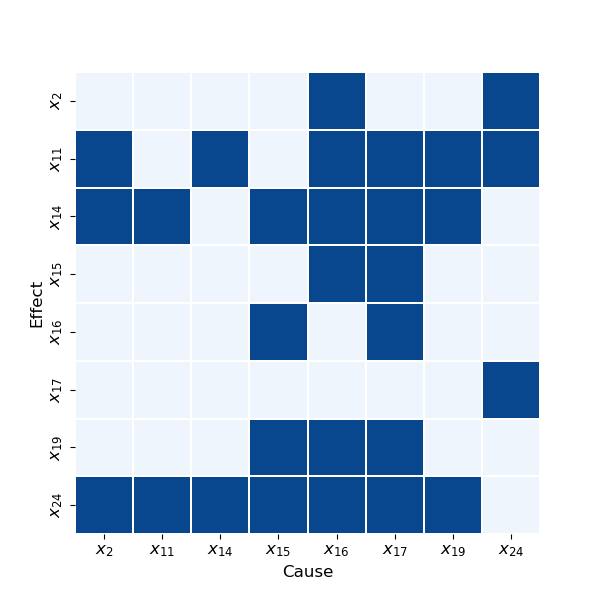} \label{fig:mff_s6_GC}
        }
        \subfigure[]{
        \includegraphics[width=0.45\linewidth]{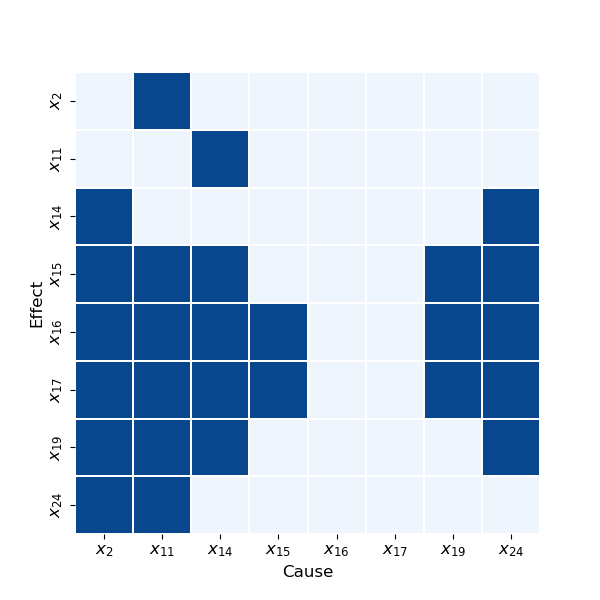} \label{fig:mff_s6_TE}
        }

    \caption{Root cause diagnosis results of different methods for fault case 6: (a) RBC; (b) GC; (c) TE.}
    \label{fig:mff_s6}
\end{figure}

In fault case 6, the 2-inch riser experienced an increase in pressure due to the opening of the bridging valve connecting the 4-inch and 2-inch risers. The bottom pressure of the 2-inch riser can be measured by the sensor PT417, and thus, the root cause variable for the fault is $x_{24}$ (Pressure in mixture zone 2-inch line).

Fig. \ref{fig:mff_s6_cont} shows the fault diagnosis results of RBC, and $x_{24}$ is identified clearly. Fig. \ref{fig:mff_s6_GC}-\ref{fig:mff_s6_TE} further illustrates the diagnostic performance of GC and TE, and 8 variables with the highest contribution in RBC results are selected as candidate variables. However, neither GC nor TE successfully identifies the correct root cause.

\begin{table}[htbp]
    \centering
	\setlength{\abovecaptionskip}{0pt}%
	\setlength{\belowcaptionskip}{10pt}%
    \caption{Top10 nodes in the root cause score rank of Root-KGD for fault case 6.}
    \label{tab:mff_s6}
    
    \begin{tabular}{cccc}
        \toprule
        Variable	&Score	&Stream and Device	&Score\\
        \midrule            
            $x_{24}$	&0.73109	&Stream 4 bottom	&0.64202\\
            $x_{1}$	&0.37281	&Stream 2	&0.39389\\
            $x_{8}$	&0.37277	&Stream 3	&0.39042\\
            $x_{15}$	&0.34912	&Stream 1	&0.36203\\
            $x_{18}$	&0.34899	&VC302	&0.35509\\
            $x_{9}$	&0.34894	&VC101	&0.32721\\
            $x_{21}$	&0.33664	&R300	&0.31865\\
            $x_{22}$	&0.31442	&PO1	&0.27083\\
            $x_{23}$	&0.25579	&T100	&0.18870\\
            $x_{19}$	&0.09780	&Stream 8	&0.16251\\
        \bottomrule
    \end{tabular}

\end{table}

Table \ref{tab:mff_s6} tabulates root cause scores of the top 10 variables (data entities in PIKG) as well as device and stream entities by the Root-KGD framework. The variable $x_{24}$ and the bottom section of stream 4 (2-inch riser) are clearly identified as the root cause, which can more effectively illustrate that the fault occurred at the bottom of the 2-inch riser and caused changes in its pressure.

\section{Conclusion}

This paper presents a novel framework for root cause diagnosis based on knowledge graphs and industrial data, called Root-KGD, and applies it to the analysis of fault root causes in industrial processes. In this framework, we represent the structural features of industrial processes through the knowledge graph, and mine the fault features in industrial data through data-driven modeling, and then combine these features to perform root cause diagnosis with the knowledge graph reasoning. In the experimental results, Root-KGD presents the ranking of root cause scores for different types of nodes and not only accurately completes the classical task of root cause variable identification but also provides interpretable support for fault root cause localization by identifying the devices and streams corresponding to the fault root causes. In addition, the framework is lightweight and real-time, capable of providing online fault root cause analysis results for industrial processes, making it more suitable for practical applications. Overall, this work represents a significant step forward in the integration of knowledge and data in the field of industrial fault root cause diagnosis.

\section*{Acknowledgements}
This work was supported in part by the National
Key Research and Development Program of China (No.
2022ZD0120003) the Key Research and Development Project of
Zhejiang Province (Grant No. 2022C01206) the National Natural
Science Foundation of China (Grant No. 61933013).


\bibliographystyle{elsarticle-num} 
\bibliography{ref}

\end{document}